\documentclass[10pt,journal]{IEEEtran}
\usepackage{enumitem,calc}
\usepackage{graphicx}
\usepackage{subfigure}
\usepackage{amsmath}
\usepackage{array}
\usepackage{subfig}
\usepackage{tabu}
\usepackage{amssymb}
\usepackage{multirow}
\usepackage{booktabs}
\usepackage{cite}
\usepackage{xcolor}
\usepackage{amsfonts}
\usepackage{makecell}
\usepackage[noend]{algpseudocode}
\usepackage{soul}
\soulregister\cite7
\soulregister\ref7
\soulregister\eqref7
\ifCLASSINFOpdf
\else
\fi
\begin{document}
%
\title{RSONet: Region-guided Selective Optimization Network for RGB-T Salient Object Detection}

\author{Bin~Wan,
	Runmin~Cong,
	Xiaofei~Zhou,
	Hao~Fang,
	Chengtao~Lv,
	and Sam~Kwong, \emph{Fellow}, \emph{IEEE}
\thanks{This work was supported in part by the the National Natural Science Foundation of China under Grant 62471278, Grant 62271180, and in part by the Research Grants Council of the Hong Kong Special Administrative Region, China under Grant STG5/E-103/24-R. \emph{(Corresponding author: Runmin Cong)}.}
	\thanks{ Bin Wan, Runmin Cong and Hao Fang are with the School of Control Science and Engineering, Shandong University, Jinan 250061, China, and also with the Key Laboratory of Machine Intelligence and System Control, Ministry of Education, Jinan 250061, China.  (E-mail:  wanbinxueshu@icloud.com; rmcong@sdu.edu.cn; fanghaook@mail.sdu.edu.cn).}
\thanks{ Xiaofei Zhou is with School of Automation, Hangzhou Dianzi University, Hangzhou 310018, China (E-mail:  zxforchid@outlook.com).}
\thanks{Chengtao lv is with School of information engineering, Huzhou University, Huzhou 313000, China (E-mail: chengtaolv@outlook.com).}

\thanks{ Sam Kwong is with the School of Data Science, Lingnan University, Tuen
	Mun, Hong Kong (E-mail: samkwong@ln.edu.hk).}
	
		
	
}

\maketitle

\begin{abstract}
	This paper focuses on the inconsistency in salient regions between RGB and thermal images. To address this issue, we propose the Region-guided Selective Optimization Network for RGB-T Salient Object Detection, which consists of the region guidance stage and saliency generation stage. In the region guidance stage, three parallel branches with same encoder-decoder structure equipped with the context interaction (CI) module and spatial-aware fusion (SF) module are designed to generate the guidance maps which are leveraged to calculate similarity scores. Then, in the saliency generation stage, the selective optimization (SO) module fuses RGB and thermal features based on the previously obtained similarity values to mitigate the impact of inconsistent distribution of salient targets between the two modalities. After that, to generate high-quality detection result, the dense detail enhancement (DDE) module which adopts the multiple dense connections and visual state space blocks is applied to low-level features for optimizing the detail information. In addition, the mutual interaction semantic (MIS) module is placed in the high-level features to dig the location cues by the mutual fusion strategy.  We conduct extensive experiments on the RGB-T dataset, and the results demonstrate that the proposed RSONet achieves competitive performance against 27 state-of-the-art SOD methods.

\end{abstract}



\IEEEpeerreviewmaketitle

\section{Introduction}
\begin{figure}[t]
	\centering
	\includegraphics[width=0.4\textwidth]{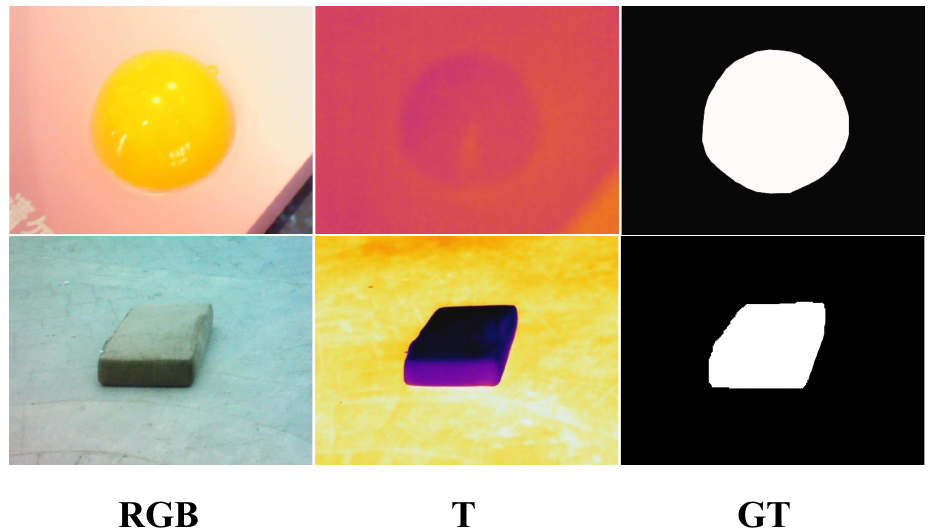}
	\caption{\small{Examples of RGB-T challenge scenarios.}}
	\label{fig_ex_t}
\end{figure}

\IEEEPARstart{S}{alient} object detection (SOD) \cite{ma2021rethinking,jing2021occlusion,li2022adjacent,zhou2024admnet,cong2025breaking,wan2025rdnet,qin2025sight,wan2026g},  as a prominent computer vision task, aims to mimic the human perceptual ability to identify the main object within a scene. Unlike traditional object detection tasks that require bounding boxes around the object area, SOD emphasizes the target area at the pixel level. Consequently, it can serve as a pre-processing technique for various scene applications, such as semantic segmentation \cite{zhou2021group,cong2022bcs,cong2024query,peng2024hsnet,cong2025divide,chen2025empowering,xiong2025mm,cong2025uis,chen2025replay}, image enhancement \cite{chen2023improving,zheng2024quad,jiang2024mutual,liu2024pyramid,cong2025reference}, person re-identification \cite{wang2023relation, yang2024stfe}, and image understanding \cite{hao2024hierarchical,xing2025towards}. In recent years, the rapid advancement of deep learning technology \cite{lian2024diving,cong2025generalized,fang2025decoupled} has significantly improved salient object detection. However, when dealing with challenging scenes such as complex backgrounds, low contrast, and unclear boundaries, RGB image-based SOD methods exhibit suboptimal performance. Therefore, it is necessary to incorporate additional information to compensate for the limitations of RGB images.  At the early stage, benefiting from the practicality of various depth information acquisition devices which provide  more shape structure and boundary information, depth cues are introduced to the SOD tasks, generating the RGB-D salient object detection \cite{wu2023mfenet,zhang2024fastersal,chen2024trans,cong2025trnet} and enhancing the detection performance to a certain extent. Nevertheless, when the object and background regions are adjacent, accurately distinguishing the object in the depth map becomes extremely challenging, thereby undermining the utility of depth information. 
To address this issue, researchers have attempted to introduce an additional source of information (thermal data), resulting in the development of RGB-T salient object detection. For example, in \cite{ma2023modal}, Ma \emph{et al}. proposed a novel modal complementary fusion network which designs the modal reweight module and spatial complementary fusion module to mitigate the impact of low-quality images. In \cite{lv2024transformer}, Lv \emph{et al}. proposed a transformer-based cross-modal integration network where the cross-modal feature fusion module and interaction-based feature decoding block are leveraged to explore the complementarity of two-modal information. In \cite{zhou2024frequency}, Zhou \emph{et al}. explored RGB-T salient object detection from the frequency perspective, and proposed the frequency-aware multi-spectral feature aggregation module and HF-guided signed distance map prediction module. Although thermal information is not affected by spatial position, it may still be influenced by other factors such as environmental conditions and material properties, resulting in an excessive information discrepancy between the two modalities. As shown in Fig. \ref{fig_ex_t}, we can see that the object area in the first row is almost indistinguishable from the background area in the thermal map. In contrast, the brick in the second row exhibits similar characteristics to the background area in the RGB image. Most existing methods fuse the two modalities using operations such as addition, multiplication, concatenation, or various attention mechanisms. However, these fusion strategies inherently assume equal importance between the modalities, which can lead to the introduction of substantial irrelevant noise when significant information disparities exist between them, thereby impairing the accurate segmentation of object regions.

To address the issues mentioned above, we propose a  Region-guided Selective Optimization Network (RSONet) for RGB-T Salient Object Detection which consists of region guidance and saliency generation phases to mitigate the negative impact stemming from the uneven distribution of the saliency regions across the two modalities and yield high-quality saliency map. 

\subsection{Region Guidance}

\begin{figure}[t]
	\centering
	\includegraphics[width=0.5\textwidth]{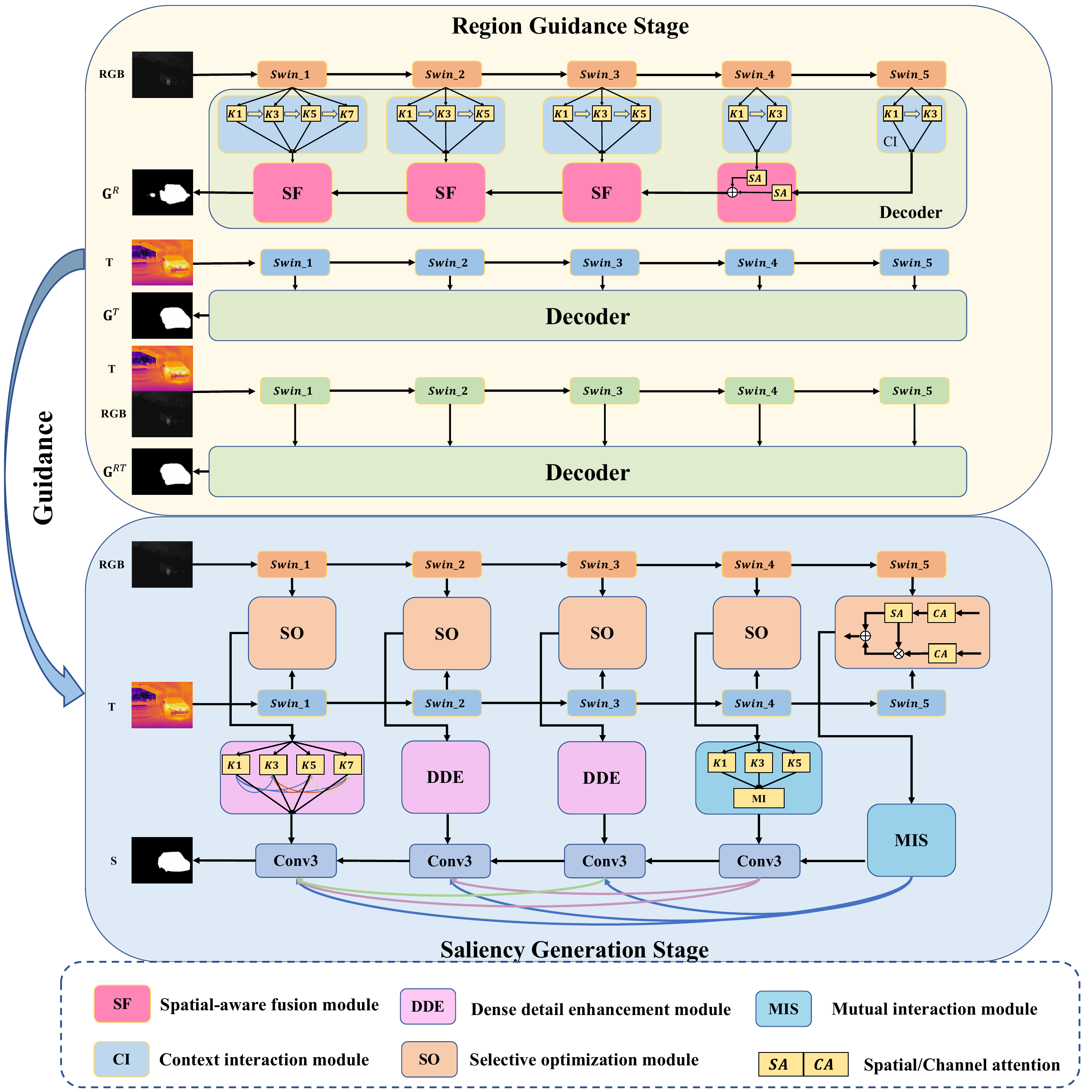}
	\caption{\small{Illustration of the RSONet.}}
	\label{fig_rg}
\end{figure}

To address the issue of inconsistent salient regions, we first design the region guidance phase to ensure that the modality with more accurate salient information takes the lead  during the subsequent bimodal fusion stage. As shown in Fig. \ref{fig_rg}, the region guidance phase contains three parallel branches which adopt the same encoder-decoder structure, where the inputs for the first two branches (\emph{i.e.,}  R and T) are the RGB image and the thermal map, respectively, while the sum of the RGB image and the thermal map serves as the input for the third branch  (\emph{i.e.,}  RT). Firstly, SwinTransformer \cite{liu2021swinnet} as the backbone network is adopted to extract multi-level features from the input. To further excavate the saliency region, we design the context interaction (CI) module, which employs convolutions with varying kernel sizes for different levels of features, enabling the capture of contextual information of the object area across different receptive fields. Then, in order to drive the generation of saliency guidance maps for three branches, the spatial-aware fusion (SF) module is proposed to fuse features yielded from the CI module layer by layer, where the spatial attention mechanism is employed to encourage the model to focus on salient object areas.  Considering that the input of branch RT is the sum of the RGB and thermal images, the resulting guidance map $\mathbf{G}^{RT}$ contains richer object information. Therefore, we compare the similarity between the guidance maps of branch R and branch T with the result of branch RT, respectively, to determine which modality should dominate during the modality fusion stage for saliency generation.

\begin{figure}[t]
	\centering
	\includegraphics[width=0.45\textwidth]{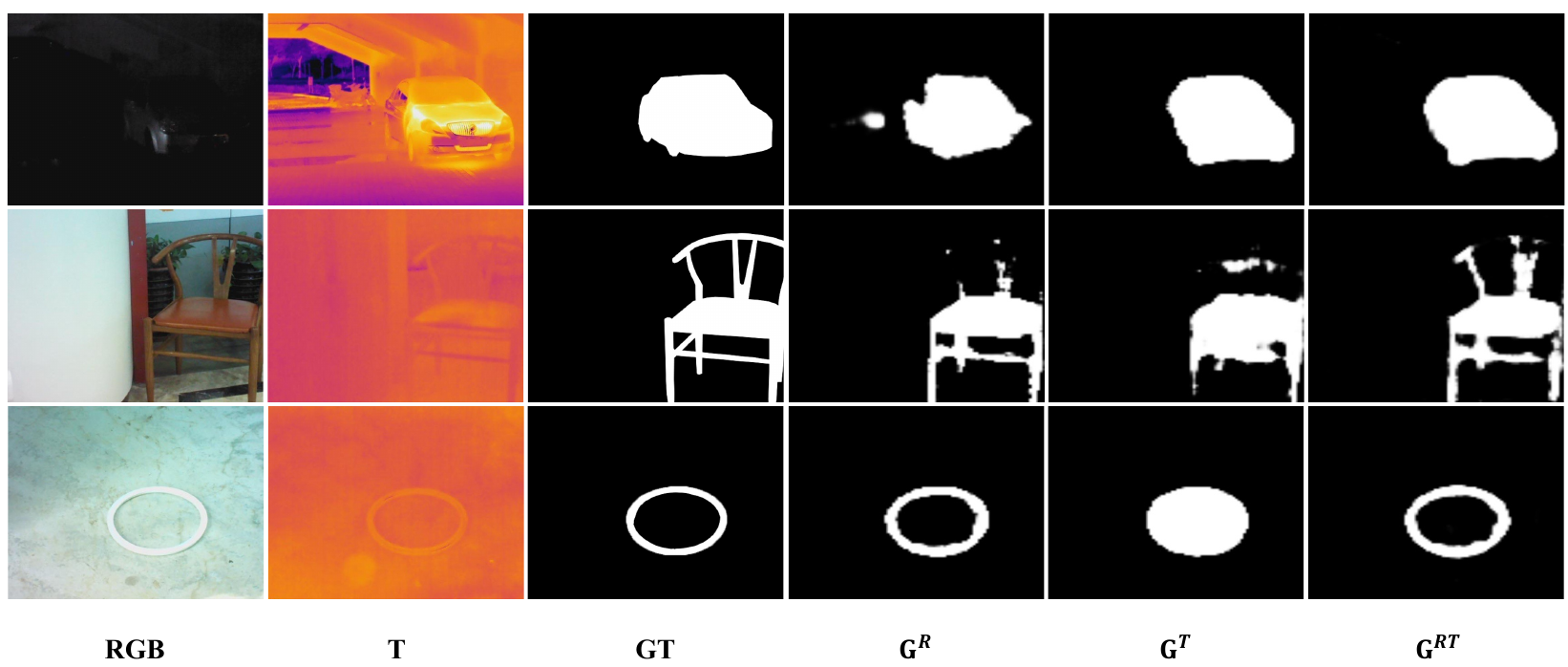}
	\caption{\small{Illustration of the guidance map.}}
	\label{fig_gm}
\end{figure}

\subsection{Saliency Generation}

After the region guidance stage, the modality with the most accurate salient information is selected, which dominates the bimodal feature fusion process in the subsequent selective optimization (SO) module to mitigate the impact caused by the uneven distribution of salient object areas. After that, considering that the inherent characteristics of multi-level features where high-level features have accurate location information, while low-level features have detailed spatial structure information, we propose the dense detail enhancement (DDE) module and mutual interaction semantic (MIS) module to extract and optimize the detail and location cues in the subsequent decoding stage. First, recognizing that object areas in natural scenes often have irregular appearances, we place the dense detail enhancement (DDE) modules in the features of the lower three layers, where the multiple dense connections are employed to facilitate the fusion of features with different receptive fields. Besides, we append four additional visual state space (VSS) blocks to further explore detailed information. Secondly, different from the DE module which leverages convolutional layers with higher dilated rates to extract structure cues, in the mutual interaction semantic (MIS) module, we just utilize convolutional layers with dilation rates of 1, 2, 3, and adopt the mutual fusion strategy to complete mutual interaction of information at different scales. Finally, cross-level connections are employed to integrate positional and spatial structural information, resulting in high-quality detection outcomes.


The main contributions of this paper are summarized as follows,
\begin{enumerate}[leftmargin=*]
\item We propose a novel Region-guided Selective Optimization Network (RSONet) for RGB-T Salient Object Detection which consists of the region guidance and saliency generation stages to resolve the issue of inconsistent salient information between the two modalities. Experimental results on  public RGB-T dataset prove that our RSONet outperforms the other state-of-the-art saliency detection methods.
\item  We propose the region guidance stage which includes three parallel branches and leverages the context interaction (CI) module and spatial fusion module to pick out the modality with the most accurate object information.
\item We propose the saliency generation stage to gain the high-quality final saliency map, where the selective optimization (SO) module is leveraged to integrate RGB and  thermal features, the dense detail enhancement (DDE) module adopts convolutions with different dilation rates to dig the spatial structure information, and the mutual interaction semantic (MIS) module is placed in the high-level features to mine the location cues.
\end{enumerate}



\section{Related Works}
In this section, we briefly review the salient object detection, including the single-modal SOD and multi-modal SOD.

\subsection{Single-modal Salient Object Detection}
In the last decade, many efforts have been spent in the one-modal \cite{li2016single} salient object detection. For example, In \cite{feng2019attentive}, Feng \emph{et al}. introduced attentive feedback modules to effectively capture object structures. In \cite{zhang2020dense}, Zhang \emph{et al}. proposed an end-to-end dense attention fluid network, utilizing a global context-aware attention module to capture long-range semantic relationships and a dense attention fluid structure to transfer shallow attention cues to deeper layers. In \cite{wang2021renet}, Wang \emph{et al}. developed a rectangular convolution pyramid and edge enhancement network, with the rectangular convolution pyramid module designed to characterize defects with diverse structures. In \cite{ma2023boosting}, Ma  \emph{et al}. proposed a bilateral extreme stripping  encoder, a dynamic complementary attention module  and a switch-path decoder to get the broader receptive fields and obtain the ability to perceive extreme large- or small-scale objects. In \cite{zhang2023salient}, Zhang \emph{et al}. addressed challenges in utilizing edge cues and multi-level feature fusion persist by a two-stream model for saliency and edge detection, an edge-guided interaction module for feature enhancement, and specialized fusion modules for progressive integration of semantic and detailed information. In \cite{wang2023elwnet}, Wang \emph{et al}. proposed a fast and extremely lightweight end-to-end wavelet neural network which designs proposing a wavelet transform module, a wavelet transform fusion module and a novel feature residual mechanism for real-time salient object detection. In \cite{yan2024asnet},  presented the adaptive semantic network Including an adaptive semantic matching module for aligning global and local contexts, an adaptive feature enhancement module for enhancing salient regions and restoring resolution, and a multiscale fine-grained inference module for refining high-level semantics with low-level details to produce high-quality saliency maps.
\subsection{Multi-modal Salient Object Detection}
With advancements in image acquisition technology, depth modality has become widely utilized in salient object detection (SOD) tasks, offering complementary information to enhance detection accuracy. In \cite{huang2022middle}, Huang \emph{et al}. proposed a lightweight RGB-D SOD model that uses two shallow subnetworks to extract low- and middle-level features from unimodal RGB and depth inputs, ensuring efficient and accurate feature representation. Similarly, in \cite{song2022improving}, Song \emph{et al}. introduced a modality-aware decoder designed to enhance multimodal integration through a series of feature embedding, modality reasoning, and feature back-projection and collection strategies, effectively improving the model's ability to handle complex scenes. In \cite{fang2024grouptransnet}, Fang \emph{et al}. proposed a novel group transformer network capable of capturing long-range dependencies across layer features, enhancing feature representation between high-level and low-level features. In \cite{feng2024mfur}, Feng \emph{et al}. introduced a multimodal multilevel feature fusion module for enhanced RGB-D features, a multi-input aggregation module to integrate RGB and depth streams, and a saliency refinement module to eliminate redundancy and refine features before decoder integration.
Depth information in RGB-D-based SOD methods is often influenced by the object's location, making it challenging to detect targets close to the background. To address this limitation, thermal images have been introduced as supplementary information for SOD. In \cite{chen2022cgmdrnet}, Chen \emph{et al}. proposed a Cross-Guided Modality Difference Reduction Network that employs a modality difference reduction module, a cross-attention fusion module, and a transformer-based feature enhancement module to minimize modality differences. Similarly, in \cite{he2022eaf}, He \emph{et al}. introduced an Enhancement and Aggregation Feedback Network, incorporating a feature enhancement block and a cross-feature aggregation module to achieve effective multimodal complementation. In \cite{wang2024intra}, Wang \emph{et al}. proposed the intra-modality self-enhancement mirror network which presents the intra-modality cross-scale self-enhancement module to exploit saliency clues by modeling the correlation between intra-modality cross-scale features. In \cite{yue2024salient}, Yue \emph{et al}. proposed spatial-frequency feature exploration modules and spatial-frequency feature interaction modules to separate spatial-frequency feature and integrates cross-modality and cross-domain information  for generating high-quality saliency maps. Furthermore, in \cite{he2025samba}, He \emph{et al}. proposed a unified pure mamba-based framework,  which redesigned the scanning strategy and upsampling process to preserve the spatial continuity of salient regions and enhance hierarchical feature aggregation for general SOD tasks. In \cite{liu2026samsod}, Liu \emph{et al}. proposed a SAM-based SOD method, which addressed modality imbalance and gradient conflicts in RGB-T SOD by incorporating unimodal supervision, gradient deconfliction, and decoupled adapters to enhance foreground?background discrimination and improve model convergence and generalization.

\section{Proposed Framework}
\subsection{Overview of Proposed RSONet}
In this paper, we propose a  Region-guided Selective Optimization Network (RSONet) for RGB-T Salient Object Detection, which comprises two stages: region guidance and saliency generation, to address the issue of inconsistent salient regions and yield high-quality saliency map. In the region guidance stage, three identical encoder-decoder structures are used to construct guidance maps $\mathbf{G}^{R}$, $\mathbf{G}^{T}$ and $\mathbf{G}^{RT}$  from the RGB image, the thermal image, and the sum of the RGB and thermal images, respectively, where the encoder adopts SwinTransformer \cite{liu2021swinnet} to extract multi-level features  $\{\mathbf{F}_i^{R}\}_{i=1}^5$ ,  $\{\mathbf{F}_i^{T}\}_{i=1}^5$ and  $\{\mathbf{F}_i^{RT}\}_{i=1}^5$ from inputs, and the decoder first leverages  convolutions of different kernel sizes in the context interaction (CI) module to extract contextual information $\{\mathbf{F}_i^{RC}\}_{i=1}^5$, $\{\mathbf{F}_i^{TC}\}_{i=1}^5$ and $\{\mathbf{F}_i^{RTC}\}_{i=1}^5$ from features at different levels. Then, the spatial-aware fusion (SF) module is designed to optimize context features, yielding the guidance maps. Next, $\mathbf{G}^{R}$ and $\mathbf{G}^{T}$ are each compared with $\mathbf{G}^{RT}$ to calculate similarity, thereby selecting the modality with the most accurate object information. In the saliency generation stage, based on the previous similarity calculation results, the selective optimization (SO) module selectively optimizes and fuses the two modality features $\{\mathbf{F}_i^{R}\}_{i=1}^5$  and $\{\mathbf{F}_i^{T}\}_{i=1}^5$  extracted from the backbone network. Subsequently, considering the different characteristics of multi-level features, the dense detail enhancement (DDE) module and mutual interaction semantic (MIS) module are placed in low-level and high-level features, respectively, which  employ different dilated convolution strategies to extract spatial structural information $\mathbf{F}_i^{D}$  and positional information $\mathbf{F}_i^{M}$ of the object. Finally, the cross-level connection is adopted to generate high-quality saliency map $\mathbf{S}$

\subsection{Region Guidance Stage}
Lighting and temperature variations often result in incomplete or unclear representations of the object region in both RGB and thermal images. Consequently, directly fusing the two modalities may introduce considerable noise from irrelevant regions, ultimately compromising detection performance. To mitigate this issue, we propose the region guidance stage, which consists of three identical encoder-decoder structures used to extract region guidance maps from the RGB image, the thermal image, and the sum of the RGB and thermal images, respectively, to select the modality with the most accurate object information and ensures it dominates the subsequent bimodal fusion stage.

\begin{figure}[t]
	\centering
	\includegraphics[width=0.45\textwidth]{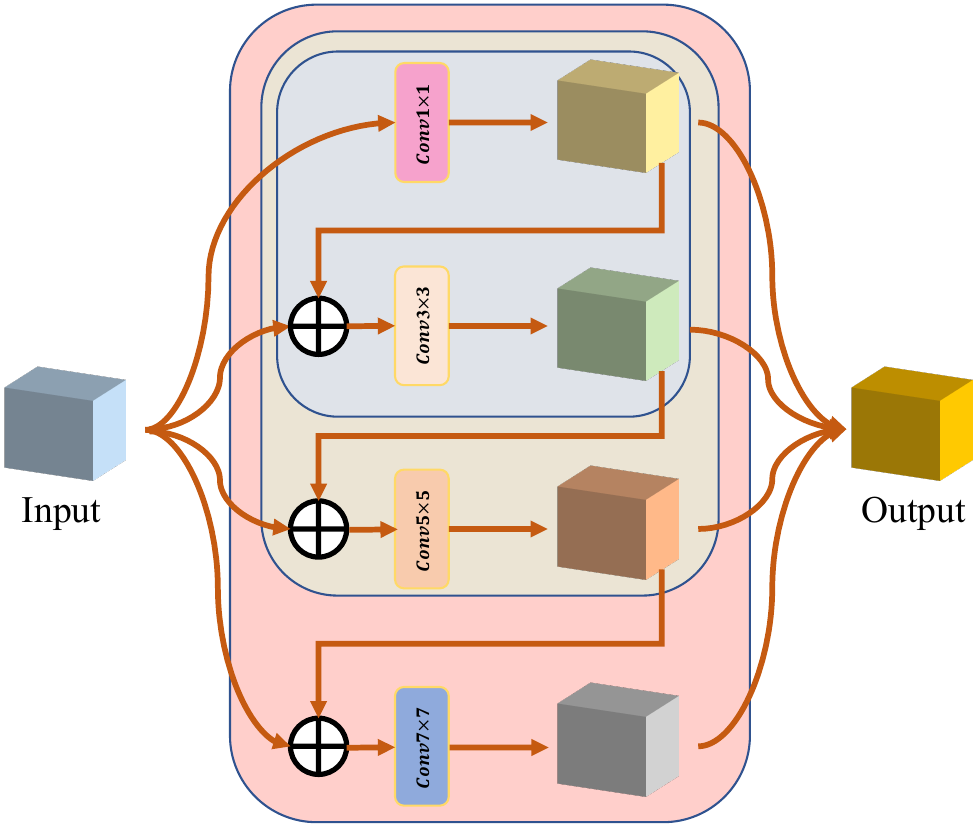}
	\caption{\small{Architecture of the context interaction module.}}
	\label{fig_ci}
\end{figure}

\subsubsection{Context Interaction Module}
Taking the RGB branch as an example,  to extract object contextual information from the multi-level features obtained by the backbone network (Swin Transformer). Considering that features at different levels exhibit distinct characteristics$-$high-level features have smaller spatial resolutions and capture object semantics and location, while low-level features retain larger spatial resolutions and preserve fine-grained structural details$-$uniformly optimizing or enhancing all levels of features during the decoding stage may obscure their unique contributions. For example, when large convolutional kernels are applied to low-resolution features, they may integrate the target and surrounding background information indiscriminately. Conversely, when small kernels are applied to high-resolution features, the receptive field may be insufficient to capture the complete structure of the target object. To end this, we leverage convolutional layers with different kernel sizes  and employ different contextual information extraction strategies for different levels of features. As shown in Fig. \ref{fig_ci}, the context interaction module consists of three variants (\emph{i.e.,} pink, yellow and blue parts), which are applied to low-level feature $\mathbf{F}_1^{R}$, mid-level features $\mathbf{F}_{2/3}^{R}$, and high-level features $\mathbf{F}_{4/5}^{R}$, respectively. For the variant one (pink part),  feature $\mathbf{F}_1^{R}$ is fed into  four parallel branches, each with different convolutional layers to extract contextual information at various scales. Besides, break down the barriers between features of different scales, we add the output of the previous branch to the input features of the current branch. Then, the results of the four branches are concatenated along the channel dimension to obtain the module's output $\mathbf{F}_1^{RCI}$,

\begin{equation}
	\begin{cases}
		\mathbf{F}_1^{R_1} =f_{1\times1}(\mathbf{F}_1^{R}) \\
		
		\mathbf{F}_1^{R_2} =f_{3\times3}(\mathbf{F}_1^{R_1} +\mathbf{F}_1^{R} )\\
		
		\mathbf{F}_1^{R_3} =f_{5\times5}(\mathbf{F}_1^{R_2} +\mathbf{F}_1^{R})\\
		
		\mathbf{F}_1^{R_4} =f_{7\times7}(\mathbf{F}_1^{R_3} +\mathbf{F}_1^{R})
	\end{cases},
\end{equation}
where $f_{1\times1}$, $f_{3\times3}$, $f_{5\times5}$ and $f_{7\times7}$ mean the convolutional layers with kernel sizes of 1, 3, 5, and 7, respectively.  For the mid-level features $\mathbf{F}_{2/3}^{R}$, considering their reduced resolution, using excessively large convolutional kernels would integrate a significant amount of irrelevant information. Therefore, compared to variant one, we remove the $7\times7$ convolutional branch,

\begin{equation}
	\begin{cases}
		\mathbf{F}_{2/3}^{R_1} =f_{1\times1}(\mathbf{F}_{2/3}^{R}) \\
		
		\mathbf{F}_{2/3}^{R_2} =f_{3\times3}(\mathbf{F}_{2/3}^{R_1} +\mathbf{F}_{2/3}^{R} )\\
		
		\mathbf{F}_{2/3}^{R_3} =f_{5\times5}(\mathbf{F}_{2/3}^{R_2} +\mathbf{F}_{2/3}^{R})\\
	\end{cases}. 
\end{equation}

Similarly, for the high-level featues $\mathbf{F}_{4/5}^{R}$, building on the modifications in variant two, we further remove the $5\times5$ convolutional branch, 
\begin{equation}
	\begin{cases}
		\mathbf{F}_{4/5}^{R_1} =f_{1\times1}(\mathbf{F}_{4/5}^{R}) \\
		
		\mathbf{F}_{4/5}^{R_2} =f_{3\times3}(\mathbf{F}_{4/5}^{R_1} +\mathbf{F}_{4/5}^{R} )
	\end{cases}. 
\end{equation}

\subsubsection{Spatial-aware Fusion Module}
\begin{figure}[!t]
	\centering
	\includegraphics[width=0.45\textwidth]{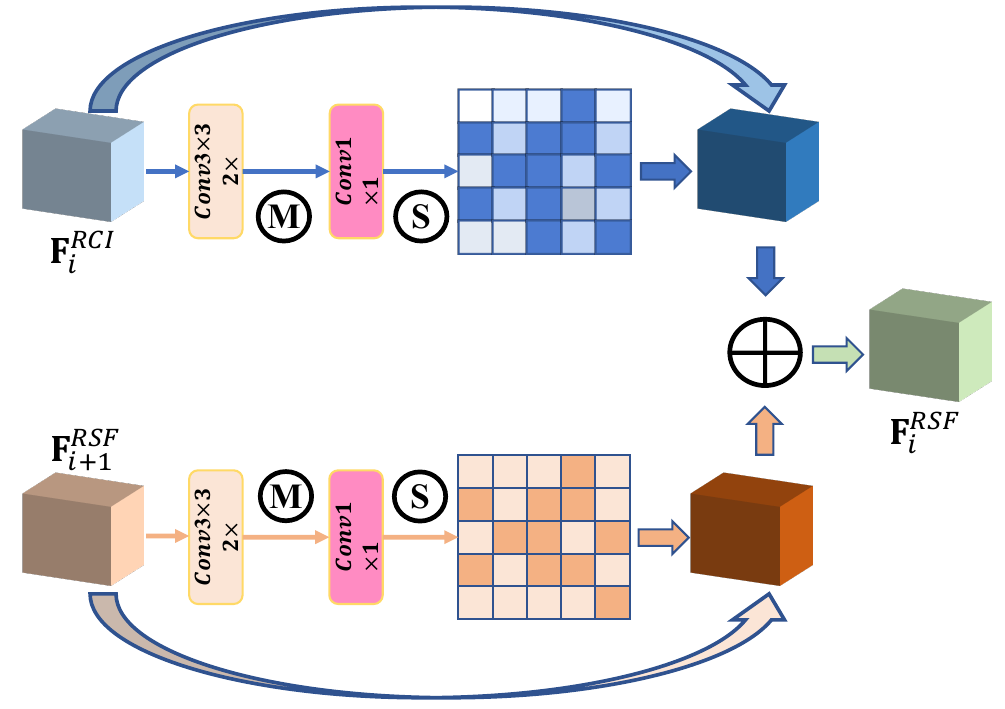}
	\caption{\small{Architecture of the spatial-aware fusion module.}} 
	\label{fig_sf}
\end{figure}
After the context interaction module, rich object information is digged out. However, the use of multiple convolutional layers of different sizes inevitably introduces irrelevant information, making it difficult to accurately reconstruct the object area. To address the issue, the spatial-aware fusion (SF) module is proposed to integrate features generated from the CI module, where the object cures are optimized in the spatial dimension. As shown in Fig. \ref{fig_sf}, first, two $3\times3$ convolutional layers are deployed to the context feature $\mathbf{F}_i^{RCI}$ to further extract object information, obtaining feature $\mathbf{F}_i^{RC}$. Then, to achieve optimization in the spatial dimension, the global max pooling followed by $1\times1$ convolutional layer and sigmoid function is applied to generate the weighted feature $\mathbf{W}_i^{R}$ which is multiplied and added with  $\mathbf{F}_i^{RC}$ to achieve feature optimization,
\begin{equation}
	\begin{cases}
		\mathbf{F}_{i}^{RC} = f_{3\times3}(f_{3\times3}(\mathbf{F}_{i}^{RCI}))\\
		
		\mathbf{W}_i^{R} = \sigma(f_{1\times1}(f_{Max}(\mathbf{F}_{i}^{RC} )))\\
		
		\mathbf{F}_{i}^{RO} = 	\mathbf{F}_{i}^{RC} \times \mathbf{W}_i^{R}  +\mathbf{F}_{i}^{RC}\\
	\end{cases}. 
\end{equation}

For the feature $\mathbf{F}_{i+1}^{RSF}$ yielded from the previous SF module, the same operations performed on feature $\mathbf{F}_i^{RCI}$ are also carried out on feature $\mathbf{F}_{i+1}^{RSF}$. Finally, two optimized features are combined through addition to produce the output feature $\mathbf{F}_{i}^{RSF}$,

\begin{equation}
	\begin{cases}
		\mathbf{F}_{i+1}^{RC} = f_{3\times3}(f_{3\times3}(\mathbf{F}_{i+1}^{RSF}))\\
		
		\mathbf{W}_{i+1}^{R} = \sigma(f_{1\times1}(f_{Max}(\mathbf{F}_{i+1}^{RC} )))\\
		
		\mathbf{F}_{i+1}^{RO} = 	\mathbf{F}_{i+1}^{RC} \times \mathbf{W}_{i+1}^{R}  +\mathbf{F}_{i+1}^{RC}\\
		
		\mathbf{F}_{i}^{RSF} = \mathbf{F}_{i}^{RO} + \mathbf{F}_{i+1}^{RO}\\
	\end{cases},
\end{equation}
where $f_{Max}$ means the global max pooling and $ \sigma$ denotes the sigmoid function.
\subsubsection{Similarity calculation}
To select the modality containing accurate object information, we first leverage $1\times1$ convolutional layer and sigmoid function to generate three guidance maps $\mathbf{G}^{R}$, $\mathbf{G}^{T}$ and $\mathbf{G}^{RT}$,

\begin{equation}
	\begin{cases}
		\mathbf{G}^{R} = \sigma(f_{1\times1}(\mathbf{F}_{1}^{RSF}))\\
		
		\mathbf{G}^{T} = \sigma(f_{1\times1}(\mathbf{F}_{1}^{TSF}))\\
		
		\mathbf{G}^{RT} = \sigma(f_{1\times1}(\mathbf{F}_{1}^{RTSF}))\\
		
	\end{cases}.
\end{equation}

Then, we compute the sum of all elements in each of the three guidance maps, obtaining three values $M^R$, $M^T$ and $M^{RT}$. After that, $M^R$ and $M^T$  are each compared with $M^{RT}$, and the smaller the difference, the more accurate the object information in the corresponding modality,
\begin{equation}
	\begin{cases}
		M^R=\frac{1}{W\times H}{\textstyle \sum_{x=1}^{W}}{\textstyle \sum_{y=1}^{H}}\mathbf{G}^{R}(x,y)\\
		
		M^T=\frac{1}{W\times H}{\textstyle \sum_{x=1}^{W}}{\textstyle \sum_{y=1}^{H}}\mathbf{G}^{T}(x,y)\\
		
		M^{RT}=\frac{1}{W\times H}{\textstyle \sum_{x=1}^{W}}{\textstyle \sum_{y=1}^{H}}\mathbf{G}^{RT}(x,y)\\
		
	\end{cases}.
\end{equation}
\subsection{Saliency Generation Stage}
Following the region guidance stage, which selects the modality with more reliable object information, we design a series of modules$-$namely, the selective optimization (SO) module, dense detail enhancement (DDE) module, and mutual interaction semantic (MIS) module$-$to effectively exploit and integrate complementary bimodal features, thereby facilitating accurate and robust saliency prediction.

\subsubsection{Selective Optimization Module}
\begin{figure}[t]
	\centering
	\includegraphics[width=0.4\textwidth]{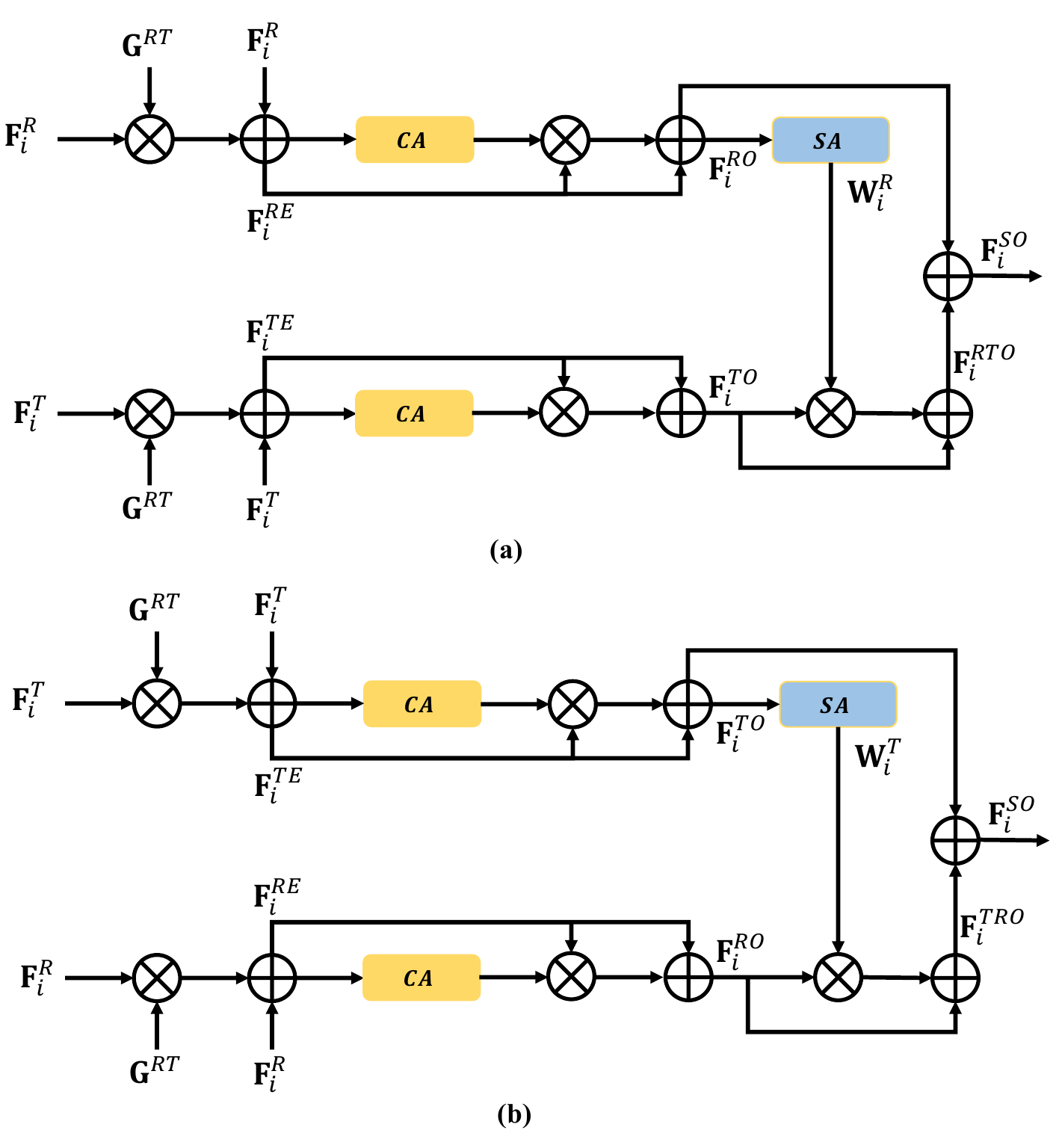}
	\caption{\small{Architecture of the selective optimization module.}}
	\label{fig_so}
\end{figure}
 As shown in Fig. \ref{fig_so} (a), where the RGB image contains richer object information compared to the thermal image. First, RGB feature $\mathbf{F}_i^{R}$ and thermal feature $\mathbf{F}_i^{T}$ are each multiplied and added with guidance map $\mathbf{G}^{RT}$ to perform an initial optimization and enhancement of the two modality features, yielding $\mathbf{F}_i^{RE}$ and $\mathbf{F}_i^{TE}$. Then, although the introduction of guidance map $\mathbf{G}^{RT}$ enhances the bimodal features to some extent, it inevitably introduces some interference since it is not an accurate saliency map. Therefore, we apply the channel attention operation which consists  of $1\times1$ convolutional layer, global average pooling and sigmoid function to feature $\mathbf{F}_i^{RE}$ and feature $\mathbf{F}_i^{TE}$ for gaining vectors $\mathbf{V}_i^{R}$ and $\mathbf{V}_i^{T}$. After that, two feature vectors are each multiplied and added to their corresponding features to complete the channel dimension feature optimization, obtaining features $\mathbf{F}_i^{RO}$ and $\mathbf{F}_i^{TO}$,
 \begin{equation}
 	\begin{cases}
 		\mathbf{F}_i^{RE} = \mathbf{F}_i^{R} \times \mathbf{G}^{RT} + \mathbf{F}_i^{R}\\
 		
 		\mathbf{F}_i^{TE} = \mathbf{F}_i^{T} \times \mathbf{G}^{RT} + \mathbf{F}_i^{T}\\
 		
 		\mathbf{F}_i^{RO} = f_{CA}(\mathbf{F}_i^{RE})\times \mathbf{F}_i^{RE}  +\mathbf{F}_i^{RE} \\
 		
 		\mathbf{F}_i^{TO} = f_{CA}(\mathbf{F}_i^{TE})\times \mathbf{F}_i^{TE}  +\mathbf{F}_i^{TE} \\
  	\end{cases}.
\end{equation}

  Considering that the RGB image contains richer object information than the thermal image in the current context, we apply spatial attention operations to feature $\mathbf{F}_i^{RO}$. The resulting weighted feature $\mathbf{W}_i^{R}$   is then multiplied and added with feature $\mathbf{F}_i^{TO}$ to achieve the optimization of the thermal image by the RGB image and generate $\mathbf{F}_i^{RTO}$. Finally, the optimized features $\mathbf{F}_i^{RO}$ and $\mathbf{F}_i^{RTO}$ are combined through addition to produce the output $\mathbf{F}_i^{SO}$,
 \begin{equation}
 	\begin{cases}
 		
 		\mathbf{F}_i^{RTO} = f_{SA}(\mathbf{F}_i^{RO})\times \mathbf{F}_i^{TO}  +\mathbf{F}_i^{TO} \\
 		
 		\mathbf{F}_i^{SO} = \mathbf{F}_i^{RO} + \mathbf{F}_i^{RTO}\\
 		
 	\end{cases}.
 \end{equation}
 where $f_{CA}$ and $f_{SA}$ denote the channel attention and spatial attention, respectively. Additionally, in another scenario where the thermal image contains more object information than the RGB image, the two modality features are fused according to the method shown in Fig. \ref{fig_so} (b) to produce the feature $\mathbf{F}_i^{SO}$.
\subsubsection{Dense Detail Enhancement Module}
\begin{figure}[t]
	\centering
	\includegraphics[width=0.48\textwidth]{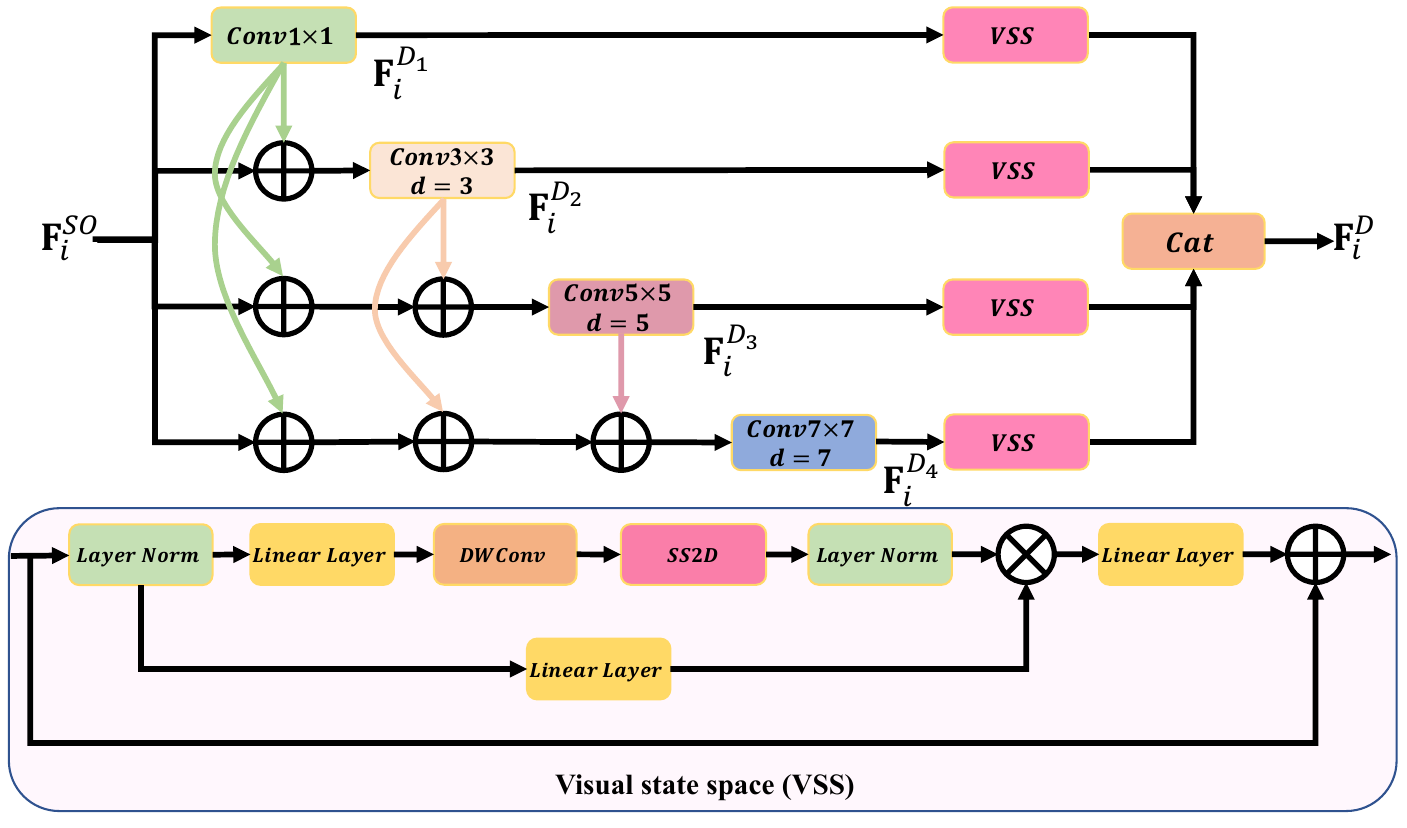}
	\caption{\small{Architecture of the dense detail enhancement module.}}
	\label{fig_dde}
\end{figure}
Considering that low-level features at high-resolution scales contain rich spatial structure information, inspired by the Atrous Spatial Pyramid Pooling (ASPP), we develop the dense detail enhancement (DDE) module. As shown in Fig. \ref{fig_dde}, the DDE module employs four parallel branches, each utilizing convolutional layers with different sizes and dilation rates to capture target information at various receptive fields. Additionally, in the ASPP, the output features from branches with different receptive fields are often fused through concatenation, which overlooks the relationships between these features. To address this, we incorporate dense connections within the DDE module, adding the receptive field features of the current branch to the input feature of the subsequent branches,
\begin{equation}
	\begin{cases}
		\mathbf{F}_i^{D_1} =f_{1\times1}(\mathbf{F}_i^{SO}) \\
		
		\mathbf{F}_i^{D_2} =f_{3\times3,3}(\mathbf{F}_i^{D_1} +\mathbf{F}_i^{SO} )\\
		
		\mathbf{F}_i^{D_3} =f_{5\times5,5}(\mathbf{F}_i^{D_1} +\mathbf{F}_i^{D_2} +\mathbf{F}_i^{SO} )\\
		
		\mathbf{F}_i^{D_4} =f_{7\times7,7}(\mathbf{F}_i^{D_1} +\mathbf{F}_i^{D_2}+\mathbf{F}_i^{D_3}  +\mathbf{F}_i^{SO})
	\end{cases}.
\end{equation}

Besides, to further capture the spatial relationship of object, we place a visual state space (VSS) block on each branch, and the output features of these blocks are concatenated along the channel dimension to obtain the output features of the DDE module,

\begin{equation}
	\mathbf{F}_i^{D} = Cat(VSS(\mathbf{F}_i^{D_1}),VSS(\mathbf{F}_i^{D_2}),VSS(\mathbf{F}_i^{D_3}),VSS(\mathbf{F}_i^{D_4})).
\end{equation}
\subsubsection{Mutual Interaction Semantic Module}
\begin{figure}[t]
	\centering
	\includegraphics[width=0.49\textwidth]{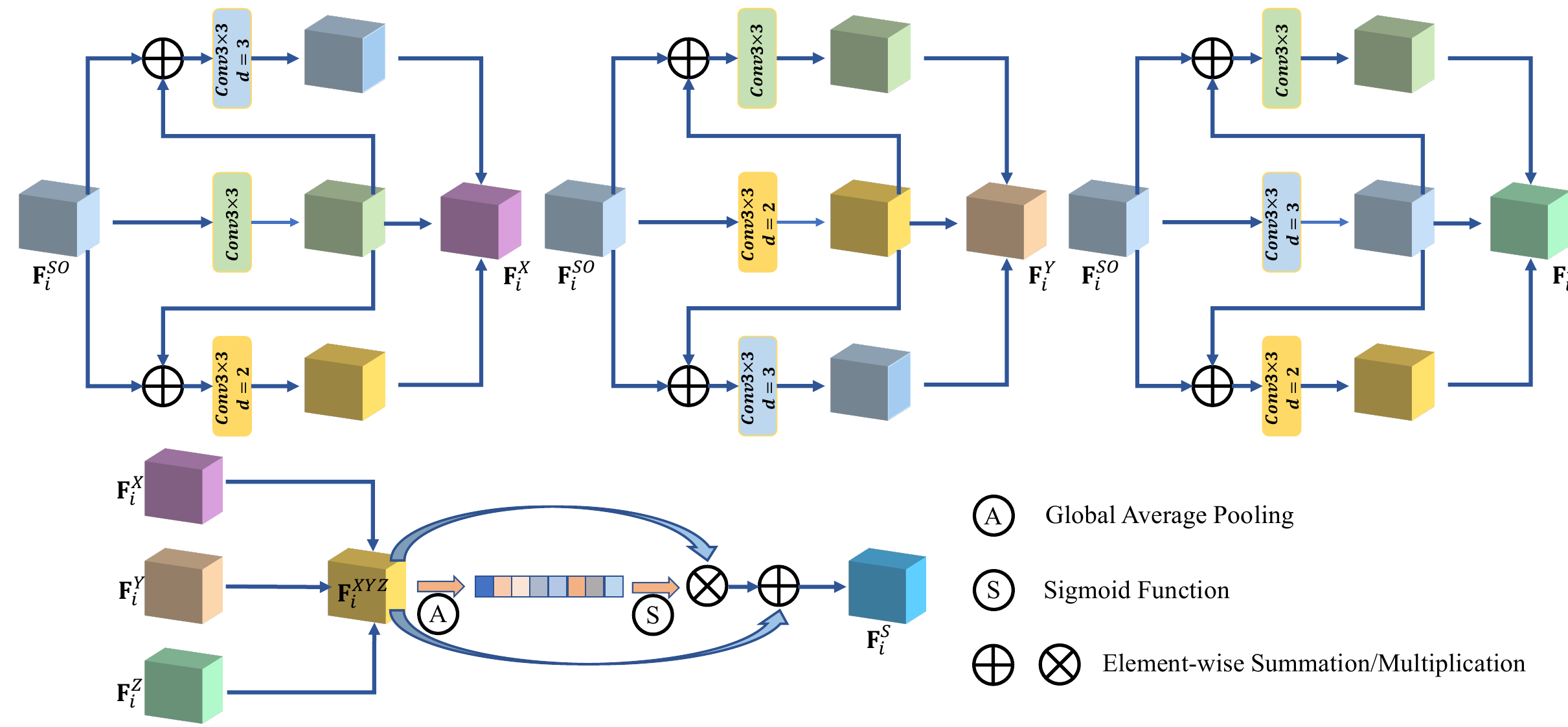}
	\caption{\small{Architecture of the mutual interaction semantic module.}}
	\label{fig_mis}
\end{figure}
In natural scenes, the positions of objects are often random, making traditional methods that use attention mechanisms ineffective in accurately locating object areas, thereby reducing detection accuracy. Therefore, for high-level features, we design the mutual interaction semantic (MIS) module to further optimize location information. As shown in Fig. \ref{fig_mis}, the MIS module employs a dense connection strategy similar to the DDE Module. However, considering the smaller resolution of high-level features, the MIS module only utilizes 3x3 convolutions with dilation rates of 1, 2, and 3. Specifically, the MIS module consists of three main branches with different fusion methods, each containing three sub-branches. Each sub-branch employs $3\times3$ convolutions with different dilation rates. To enable interaction among different receptive fields, the output of each sub-branch is added to the input features of the other two sub-branches. Then, the features of  three sub-branches are fused through concatenation to obtain features $\mathbf{F}_i^{X}$, $\mathbf{F}_i^{Y}$ and $\mathbf{F}_i^{Z}$. 
\begin{equation}
	\begin{cases}
		\mathbf{F}_i^{X_1} =f_{3\times3}(\mathbf{F}_i^{SO}) \\
		
		\mathbf{F}_i^{X_2} =  f_{3\times3,2}(\mathbf{F}_i^{X_1}+\mathbf{F}_i^{SO})\\
		
		\mathbf{F}_i^{X_3} = f_{3\times3,3}(\mathbf{F}_i^{X_1}+\mathbf{F}_i^{SO})\\
		
		\mathbf{F}_i^{X} = f_{3\times3}(Cat(\mathbf{F}_i^{X_1},\mathbf{F}_i^{X_2},\mathbf{F}_i^{X_3} ))
		
	\end{cases},
\end{equation}

\begin{equation}
	\begin{cases}
		\mathbf{F}_i^{Y_1} =f_{3\times3,2}(\mathbf{F}_i^{SO}) \\
		
		\mathbf{F}_i^{Y_2} =  f_{3\times3}(\mathbf{F}_i^{Y_1}+\mathbf{F}_i^{SO})\\
		
		\mathbf{F}_i^{Y_3} = f_{3\times3,3}(\mathbf{F}_i^{Y_1}+\mathbf{F}_i^{SO})\\
		
		\mathbf{F}_i^{Y} = f_{3\times3}(Cat(\mathbf{F}_i^{Y_1},\mathbf{F}_i^{Y_2},\mathbf{F}_i^{Y_3} ))
		
	\end{cases},
\end{equation}

\begin{equation}
	\begin{cases}
		\mathbf{F}_i^{Z_1} =f_{3\times3,3}(\mathbf{F}_i^{SO}) \\
		
		\mathbf{F}_i^{Z_2} =  f_{3\times3}(\mathbf{F}_i^{Z_1}+\mathbf{F}_i^{SO})\\
		
		\mathbf{F}_i^{Z_3} = f_{3\times3,2}(\mathbf{F}_i^{Z_1}+\mathbf{F}_i^{SO})\\
		
		\mathbf{F}_i^{Z} = f_{3\times3}(Cat(\mathbf{F}_i^{Z_1},\mathbf{F}_i^{Z_2},\mathbf{F}_i^{Z_3} ))
		
	\end{cases}.
\end{equation}

Similarly, the output features of the three main branches are concatenated and fused, with a channel attention mechanism applied to suppress irrelevant information to some extent, yielding the semantic feature $\mathbf{F}_i^{S}$,

\begin{equation}
	\begin{cases}
		\mathbf{F}_i^{XYZ} =f_{3\times3}(Cat(\mathbf{F}_i^{X},\mathbf{F}_i^{Y},\mathbf{F}_i^{Z} )) \\
		
		\mathbf{F}_i^{S} = f_{CA}(\mathbf{F}_i^{XYZ})\times \mathbf{F}_i^{XYZ} + \mathbf{F}_i^{XYZ}

	\end{cases},
\end{equation}
where $f_{3\times3}$, $f_{3\times3,2}$ and $f_{3\times3,3}$ mean the convolutional layers with dilated rates of 1, 2, and 3, respectively. 
\subsection{Deep Supervision}
To promote the performance of RSONet, we adopt the fusion loss, which consists of binary cross entropy (BCE) loss\cite{de2005tutorial}, boundary intersection over union (IOU) loss\cite{rahman2016optimizing} and F-measure (FM) loss\cite{zhao2019optimizing} to supervise five saliency  maps. The loss function $L_{total}$ is formulated as follows,
\begin{equation}
	L_{total}=\frac{1}{N} \sum_{i=1}^{N}(L_{bce}+L_{iou}+L_{fm}),
\end{equation}
where $N$ is batch size of training phase.

\begin{table*}[!t]
	\scriptsize
	\setlength\tabcolsep{4.2pt}
	\renewcommand{\arraystretch}{0.8}
	\centering
	\caption{Quantitative comparisons with 27 methods on three RGB-T datasets. The top three results in each column are marked with \textcolor{red}{red}, \textcolor{blue}{blue}, \textcolor{green}{green} color, respectively.  $\uparrow$ and $\downarrow$ denote that larger value is better and smaller value is better, respectively.}
	\begin{tabular}{ccc|cccc|cccc|cccc}
		\toprule
		\multicolumn{1}{c}{}&\multicolumn{1}{c}{\multirow{2}{*}{Methods}} &\multicolumn{1}{c}{}&\multicolumn{4}{c}{VT5000} &\multicolumn{4}{c}{VT1000} &\multicolumn{4}{c}{VT821}\\
		\cline{4-15}\specialrule{0em}{1pt}{1pt}
		\multicolumn{3}{c}{}&$\mathcal{M}$ $\downarrow$& $F_{\beta}$ $\uparrow$& $S_{\alpha}$ $\uparrow$& \multicolumn{1}{c}{$E_{\xi}$ $\uparrow$}& $\mathcal{M}$ $\downarrow$& $F_{\beta}$ $\uparrow$& $S_{\alpha}$ $\uparrow$& \multicolumn{1}{c}{$E_{\xi}$ $\uparrow$}& $\mathcal{M}$ $\downarrow$& $F_{\beta}$ $\uparrow$& $S_{\alpha}$ $\uparrow$& \multicolumn{1}{c}{$E_{\xi}$ $\uparrow$} \\
		
		
		
		
		
		
		\midrule
		\multirow{16}{*}{RGB-T}

		&MIDD \cite{tu2021multi}	&TIP${}_{21}$
		&.046	&.788	&.856	&.893	&.029	&.870	&.907	&.935	&.045	&.803	&.871	&.897\\
		&MMNet \cite{gao2021unified}	&TCSVT${}_{21}$
		&.044	&.777	&.863	&.888	&.027	&.859	&.914	&.931	&.040	&.792	&.874	&.893\\
		&CSRNet \cite{huo2021efficient}	&TCSVT${}_{21}$
		&.042	&.809	&.868	&.907	&.024	&.875	&.918	&.939	&.038	&.829	&.885	&.912\\
		
		&ECFFNet \cite{zhou2021ecffnet}	&TCSVT${}_{21}$&.038	&.803	&.875	&.911	&.022	&.873	&.924	&.947	&.035	&.807	&.877	&.907\\
		
		&APNet \cite{zhou2021apnet}	&TETCI${}_{21}$
		&.035	&.816	&.876	&.917	&.022	&.880	&.922	&.950	&.034	&.814	&.868	&.911\\
		&CGFNet \cite{wang2021cgfnet}	&TCSVT${}_{21}$
		&.035	&.851	&.883	&.924	&.024	&.901	&.921	&.953	&.036	&.842	&.879	&.915\\
		
		&ADF \cite{tu2022rgbt}&TMM${}_{22}$&.048	&.778	&.864	&.891	&.034	&.846	&.909	&.922	&.077	&.716	&.810	&.844\\
		&TNet \cite{cong2022does}	&TMM${}_{22}$
		&.033	&.845	&.895	&.929	&.021	&.887	&.929	&.952	&.030	&.840	&.899	&.924\\

		&CCFENet \cite{liao2022cross}	&TCSVT${}_{22}$
		&.031	&.858	&.896	&.937	&.018	&.904	&.934	&.962	&.027	&.856	&.900	&.933\\

		&SwinNet \cite{liu2021swinnet}	&TCSVT${}_{21}$
		&.026	&.864	&.912	&.945	&.018	&.894	&.938	&.957	&.030	&.846	&.904	&.928\\
		
		&HRTransNet \cite{tang2022hrtransnet}	&TCSVT${}_{22}$
		&{.025}	&.869	&.913	&{.948}	&.017	&.898	&.938	&.956	&.026	&.851	&.906	&.933\\
		
		&UidefNet \cite{wang2022unidirectional}	&EAAI${}_{22}$&.025	&{.874}	&{.914}	&{.948}	&\textcolor{green}{.015}	&.911	&.941	&{.969}	&{.023}	&{.872}	&{.917}	&{.945}\\
		
		&MCFNet \cite{ma2023modal}&AI${}_{23}$	& .033 & .846 & .888 & .928 & .019 & .900   & .932 & .960  & .029 & .842 & .891 & .925 \\
		
		&TAGFNet \cite{wang2023thermal}&EAAI${}_{23}$& .036 & .826 & .884 & .916 & .021 & .888 & .926 & .951 & .035 & .821 & .881 & .909 \\
		
		&CAVER \cite{pang2023caver}	&TIP${}_{23}$
		&.029	&.854	&.900	&.940	&{.017}&.904	&.938	&.965	&.027	&.853	&.898	&.934
		\\
		
		&WaveNet \cite{zhou2023wavenet}	&TIP${}_{23}$
		&.026	&.863	&.912	&.944	&\textcolor{green}{.015}	&.905	&{.945}	&\textcolor{green}{.967	}&{.025}	&.854	&.912	&{.935}
		\\
		
		&SPNet \cite{zhang2023saliency} &ACM MM${}_{23}$	&.024&.885&.915 & .952 & \textcolor{green}{.015} & .918& .942 &\textcolor{blue}{ .972} & .023&.874 & .914& .943 \\

		
		&MMIR \cite{wu2024rgb}	&TIM${}_{24}$
		&.031	&.879	&.889	&.936	&.019	&\textcolor{blue}{.927}&.938	&.955	&.030	&.866	&.885	&.923\\
		
		&PATNet \cite{jiang2024patnet}	&KBS${}_{24}$
		&.023 	&.880	&.916&.951&\textcolor{green}{.015}	&.908	&.941	&.964	&.024	&.868	&.914	&.938
		\\

		&LAFB \cite{wang2024learning}&TCSVT${}_{24}$&.030&.856&.893&.934&.018&.903&.932&.960&.034&.841&.884&.922\\
		&DFENet \cite{lyu2024deep}&arXiv${}_{24}$&.023&.882&.920&.950&\textcolor{green}{.015}&.901& \textcolor{blue}{.948}&.958&.023&.871&.917&.938\\
		&ContriNet \cite{tang2024divide}&TPAMI${}_{25}$&\textcolor{green}{.022}&.878&.915&.940&\textcolor{green}{.015}&.918&.941&.954&\textcolor{green}{.020}&\textcolor{red}{.898}&\textcolor{blue}{.923}&\textcolor{red}{.956}\\
		
		&Samba \cite{he2025samba}&CVPR${}_{25}$&\textcolor{blue}{.021}&\textcolor{green}{.894}&\textcolor{red}{.926}&\textcolor{green}{.958}&\textcolor{red}{.013}&\textcolor{green}{.926}&\textcolor{red}{.952}&\textcolor{red}{.976}&\textcolor{red}{.178}&\textcolor{green}{.894}&\textcolor{red}{.934}&\textcolor{blue}{.955}\\
		
		&SAMSOD \cite{liu2026samsod}&TMM${}_{26}$&\textcolor{blue}{.021}&\textcolor{red}{.921}&\textcolor{blue}{.923}&\textcolor{red}{.964}&\textcolor{green}{.015}&\textcolor{red}{.942}&.942&\textcolor{red}{.976}&\textcolor{green}{.022}&\textcolor{blue}{.895}&.913&\textcolor{green}{.950}\\
		&DSCDNet \cite{yu2024dual}&TCE${}_{25}$&.023&.888&\textcolor{green}{.918}&.949&\textcolor{blue}{.014}&.921&\textcolor{green}{.946}&.955&\textcolor{green}{ .022}&.876&.904&.915\\
		
		&SMRNet \cite{xiao2025smr}	&AAAI${}_{25}$
		&.030	&.859	&.891	&.935	&.020	&899	&.924	&.945	&.030	&.844	&.888	&.920
		\\
		&ISMNet \cite{wang2024intra}	&TCSVT${}_{25}$
		&.025	&.885	&.913	&.945	&\textcolor{blue}{.014}	&.922	&.942	&.954	&\textcolor{blue}{.021}&.886	&.917	&.945
		\\
		
		
		&\textbf{Our}&
		&\textbf{\textcolor{red}{.020}}	&\textbf{\textcolor{blue}{.910}}	&\textbf{\textcolor{red}{.926}}	&\textbf{\textcolor{blue}{.963}}	&\textbf{\textcolor{blue}{.014}}	&\textbf{.923}	&\textbf{\textcolor{green}{.946}}	&\textbf{\textcolor{blue}{.972}}	&\textcolor{blue}{\textbf{{.021}}}	&\textbf{{.883}}	&\textbf{\textcolor{green}{.921}}	&\textbf{{.946}}\\
		\bottomrule
	\end{tabular}
	\label{tab_comparison}
\end{table*}
\section{Experiments And Analyses}
\subsection{Implementation Details and Evaluation Metrics}
\subsubsection{Implementation Details}
We train and test our RSONet on three public RGB-T datasets, including VT5000 \cite{tu2022rgbt}, VT1000 \cite{tu2019rgb}, and VT821 \cite{wang2018rgb}. The VT5000 is a large-scale RGB-T dataset and contains 5000 images of all day scenes, of which 2500 images are the training dataset and the other 2500 images are the test dataset. VT1000 and VT821 includes 1000 and 821 registered RGB-T images, which are taken as the test set.  Our proposed network is conductecd based on the Pytorch toolbox, and we implement it on a computing station with a single NVIDIA GeForce RTX 4080 GPU. In this article, for the backbone network of our RSONet, we utilize the SwinTransformer model \cite{liu2021swinnet}, which has been pre-trained on the ImageNet dataset \cite{deng2009imagenet}. Other network parameters are initialized randomly.  Before the training process, training and test images are resized to 384$\times$384 for minimizing the utilization of computing resources. Besides, to optimize the training process, we select the RMSprop optimizer \cite{tieleman2012rmsprop} to minimize the loss function, where the learning rate and momentum are set 1e-4 and 0.9, respectively.


\subsubsection{Evaluation Metrics}
To evaluate the performance of our proposed RSONet, we choose five evaluation metrics, including mean absolute error ($M$)\cite{perazzi2012saliency}, F-measure ($F_{\beta}$)\cite{achanta2009frequency}, E-measure ($E_{\xi}$) \cite{fan2018enhanced}, Structure-measure ($S_{\alpha}$)\cite{fan2017structure}. 
The mathematical definitions of all metrics are depicted as follows,
%




\begin{figure*}[!t]
	\centering
	\includegraphics[width=0.75\textwidth]{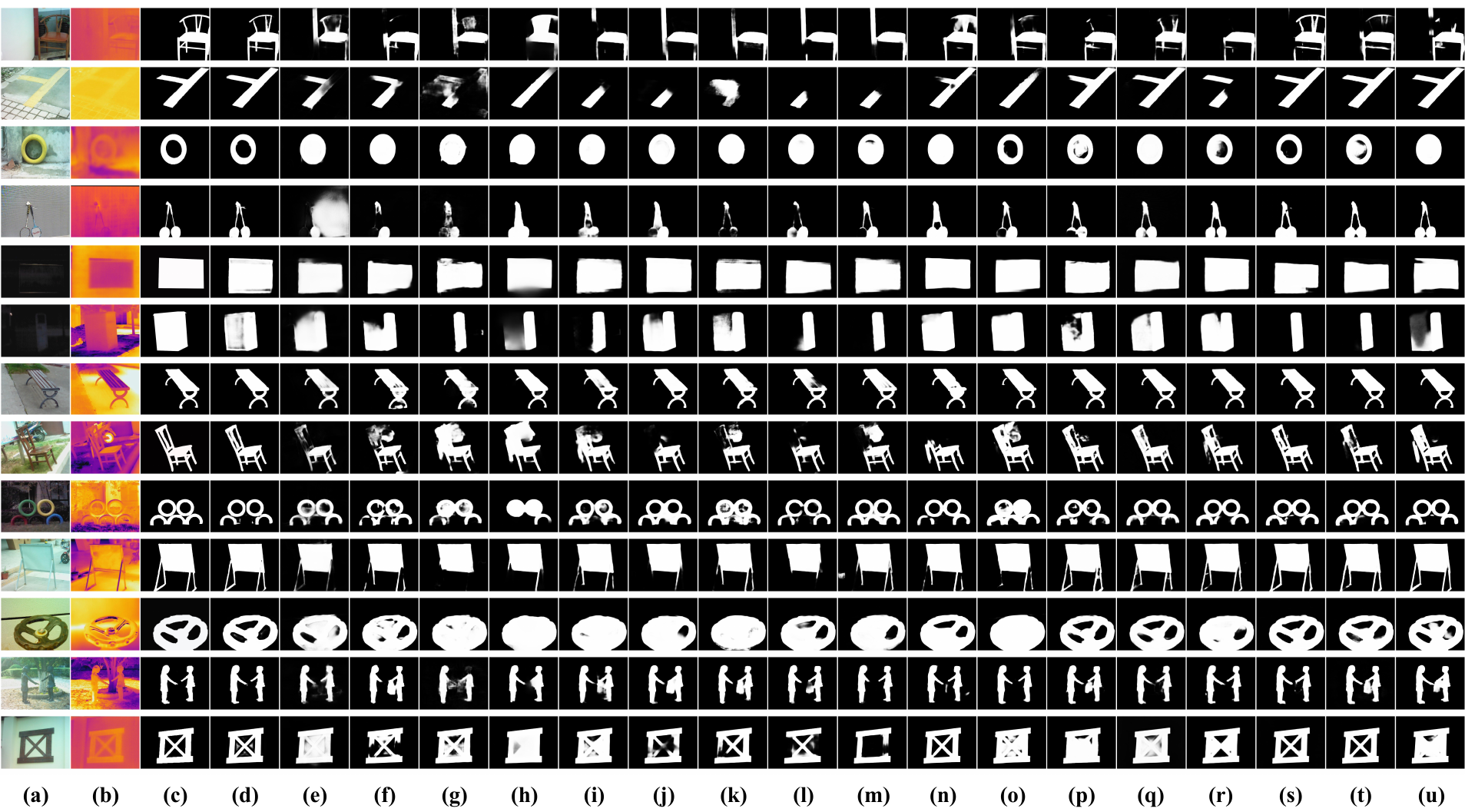}
	\caption{\small{Visual comparisons with some methods on RGB-T datasets. The predictions of compared method are listed in each column. (a) RGB images, (b) thermal images, (c) ground truth, (d) Ours, (e) ADF \cite{tu2022rgbt}}, (f) MIDD \cite{tu2021multi}, (g) MMNet \cite{gao2021unified}, (h) CSRNet \cite{huo2021efficient}	(i) ECFFNet \cite{zhou2021ecffnet}, (j) TAGFNet \cite{wang2023thermal}, (k) APNet \cite{zhou2021apnet},	(l) CGFNet \cite{wang2021cgfnet}, (m) TNet \cite{cong2022does} ,(n) MCFNet \cite{ma2023modal}, (o) CCFENet \cite{liao2022cross}, (p) CAVER \cite{pang2023caver},	(q) SwinNet \cite{liu2021swinnet}, (r) WaveNet \cite{zhou2023wavenet},	(s) HRTransNet \cite{tang2022hrtransnet},	(t) UidefNet\cite{wang2022unidirectional}, (u) SPNet \cite{zhang2023saliency}} 
	\label{fig_com}
\end{figure*}
\subsection{Comparison with State-of-the-Art Methods}
In experiments, we compare the proposed RSONet with other 27 state-of-the-art  salient object detection methods, including  ADF \cite{tu2022rgbt}, MIDD \cite{tu2021multi},
SPNet \cite{zhang2023saliency}, 
UidefNet \cite{wang2022unidirectional},
HRTransNet \cite{tang2022hrtransnet},
WaveNet \cite{zhou2023wavenet},
SwinNet \cite{liu2021swinnet},
CAVER \cite{pang2023caver},
CCFENet \cite{liao2022cross},
MCFNet \cite{ma2023modal},
TNet \cite{cong2022does},
CGFNet \cite{wang2021cgfnet},
APNet \cite{zhou2021apnet},
TAGFNet \cite{wang2023thermal},
ECFFNet \cite{zhou2021ecffnet},
CSRNet \cite{huo2021efficient},
MMIR \cite{wu2024rgb},
SMRNet \cite{xiao2025smr},
ISMNet \cite{wang2024intra},
DSCDNet \cite{yu2024dual},
DFENet \cite{lyu2024deep},
LAFB \cite{wang2024learning},
PATNet \cite{jiang2024patnet},
MMNet \cite{gao2021unified}, ContriNet \cite{tang2024divide}, Samba \cite{he2025samba} and SAMSOD\cite{liu2026samsod}. To ensure a fair comparison, we used the source code released online by the  authors.

\subsubsection{Qualitative Comparison}
To demonstrate the effectiveness  of our  RSONet, we list some visualization results of challenging scenes from RGB and thermal images, as shown in Fig. \ref{fig_com}, where the object region and background exhibit similar thermal radiation in thermal maps (rows 1-4), the object appears under low-light conditions in RGB images (rows 5-6), and the object demonstrates complex appearance contours in the images (rows 6-12).

For the challenges of similar thermal radiation, we can observe that the object region often shares similar thermal radiation characteristics with the background in the thermal images, significantly increasing the complexity and challenges of detection. Consequently, most SOD methods relying solely on RGB images frequently fail to achieve accurate segmentation of the salient object. Specifically, as illustrated in Fig. \ref{fig_com} (the $1^{st}$ row),  the chair is clearly visible in the RGB image; however, in the thermal map, the chair's thermal radiation closely resembles that of the background. From it,  we observe that nearly all comparison methods fail to fully reconstruct the various parts of the chair, with methods CAVER \cite{pang2023caver}  and WaveNet \cite{zhou2023wavenet} are able to identify only the seat surface, and methods TAGFNet \cite{wang2023thermal}, APNet \cite{zhou2021apnet}, CGFNet \cite{wang2021cgfnet} and TNet \cite{cong2022does} mistakenly identify the door frame area as a significant object. For the challenges of low-light condition, as shown in Fig. \ref{fig_com} (rows 5-6), we can see that objects are barely distinguishable in RGB images, which significantly hinders detection. For instance, in the RGB image of the $6^{th}$ row, where only the side of the box is visible, most detection methods (\emph{i.e.}, MMNet \cite{gao2021unified}, CSRNet \cite{huo2021efficient}, and HRTransNet \cite{tang2022hrtransnet}) only highlight a partial area of the object. Furthermore, detection results from other methods also exhibit missing regions in certain areas. For the other images (rows 6-12) which contain the challenges of complex appearance contours, it can be observed that various comparison methods (\emph{i.e.}, MIDD \cite{tu2021multi}, ECFFNet \cite{zhou2021ecffnet}) exhibit a certain degree of false positives or omissions. Analysis of the above visualization results reveals that our method RSONet demonstrates greater robustness and achieves superior performance across all challenging scenes.

\begin{table}[t]
	\centering

\scriptsize
	\renewcommand{\arraystretch}{1.2}
	\setlength\tabcolsep{0.3pt}
	\caption{Comparison of the model parameter, complexity and the average running speed.}
	\begin{tabular}{c c c c c c c c c c}
		\hline
		\hline
		Method&MIDD&CGFNet&TNet&SwinNet&TAGFNet&LAFB&PATNet&Samba\\
		Speed(FPS)&42.2&52.3&49.6&30.9&39.5&43&10.6&-\\
		Params(M)&52.4&69.9&87&198.8&36.2&453&94.9&54.9\\
		FLOPs(G)&216.7&347.8&39.7&124.7&115.1&139.7&51.1&71.6\\
		\hline
		Method&WaveNet&DFENet&DSCDNet&SMRNet&ISMNet&SAMSOD&ContriNet&Ours\\\
		Speed(FPS)&7.7&11.7&9.6&32.5&6.1&9.4&-&9.4\\
		Params(M)&30.2&149.6&92.3&31.5&114.3&32.7&96.3&88\\
		FLOPs(G)&26.7&139.5&134.1&45.3&100.4&418.0&126.9&143.8\\
		\hline

	\end{tabular}
	\label{tab_time}
\end{table}

\subsubsection{Quantitative comparison}  
In this part, we list the quantitative performance comparisons of our RSONet with other twenty-nine SOD methods in terms of  four evaluation metrics (\emph{i.e.}, $M$, $F_{\beta}$, $S_{\alpha}$, $E_{\xi}$), as shown in in Table. \ref{tab_comparison}, where the numerical results reveal that our  proposed method  performs better than others on RGB-T datasets (\emph{i.e.},  VT821, VT1000, and VT5000). To be specific,  
compare to state of the art method PATNet \cite{jiang2024patnet}, our RSONet achieves optimal results on VT5000 and delivers competitive performance on VT1000 and VT5000, where $F_{\beta}$, $E_{\xi}$ and $S_{\alpha}$ increase by 3.4$\%$ , 1.2$\%$ and 1.1$\%$ on VT5000, and $F_{\beta}$ and $E_{\xi}$ increase by 1.7$\%$  and 0.8$\%$ on VT1000. Furthermore, Table. \ref{tab_time}  presents the model parameters, icomplexity and detection speeds of several representative methods. As shown, the proposed model benefits from the parameter-sharing strategy, which effectively controls the number of parameters and keeps it at a moderate level compared with existing approaches. Nevertheless, due to the adoption of a two-stage design, the detection speed is relatively lower, reflecting the trade-off between structural refinement and computational efficiency.

\subsection{Ablation Study}
\begin{table}[t]
	\normalsize
	\renewcommand{\arraystretch}{0.9}
	\setlength\tabcolsep{3.8pt}
	\centering
	\caption{Ablation study on different component.}
	\begin{tabular}{c|c|c|c|c}
		\hline
		\multicolumn{1}{c|}{Settings}& \multicolumn{1}{c|}{$M$$\downarrow$}& \multicolumn{1}{c|}{$F_{\beta}$$\uparrow$}& \multicolumn{1}{c|}{$S_{\alpha}$$\uparrow$}&\multicolumn{1}{c}{$E_{\xi}$$\uparrow$}\\
		\hline
		w/o SO (Addition)&0.0217&0.8883&0.9213&0.9523\\
		w/o SO (Multiplication)&0.0208&0.8948&0.9231&0.9587\\
		w/o SO (Concatenation)&0.0215&0.8896&0.9224&0.9558\\
		w/o SO (Pixel-wise)&0.0203&0.8951&0.9239&0.9605 \\
		R$\longrightarrow$T&0.0215&0.8898&0.9230&0.9561\\
		T$\longrightarrow$R&0.0216&0.8896&0.9233&0.9554\\
		w/o DDE&0.0203&0.9082&0.9213&0.9631\\
		w/o MIS&0.0203&0.8997&0.9241&0.9593\\
		w/o DDE $\&$ MIS&0.0217&0.9053&0.8995&0.9556\\
		\textbf{Ours}&\textbf{0.0197}&\textbf{0.9071}&\textbf{0.9261}&\textbf{0.9632}\\
		\hline
		
		\hline
	\end{tabular}
	\label{tab_ablation}
\end{table}

\begin{table}[t]
	\normalsize
	\renewcommand{\arraystretch}{0.8}
	\setlength\tabcolsep{8pt}
	\centering
	\caption{Ablation study on different backbone.}
	\begin{tabular}{c|c|c|c|c}
		\hline
		\multicolumn{1}{c|}{Settings}& \multicolumn{1}{c|}{$M$$\downarrow$}& \multicolumn{1}{c|}{$F_{\beta}$$\uparrow$}& \multicolumn{1}{c|}{$S_{\alpha}$$\uparrow$}&\multicolumn{1}{c}{$E_{\xi}$$\uparrow$}
		\\
		\hline
		ResNet-18&0.0408&0.8012&0.8618&0.9052\\
		ResNet-34&0.0358&0.8146&0.8770&0.9125\\
		ResNet-50&0.0413&0.7965&0.8638&0.8994\\
		SAM&0.0438&0.8222&0.8409&0.9074\\
		DINO&0.0372&0.8562&0.8751&0.9258\\
		\textbf{Ours}&\textbf{0.0197}&\textbf{0.9071}&\textbf{0.9261}&\textbf{0.9632}\\
		\hline
		
		\hline
	\end{tabular}
	\label{tab_back}
\end{table}

		

To show the effectiveness of each component in our RSONet (\emph{i.e.}, SO, DDE and MIS), and the different role of loss function and backbone, we conduct several ablation experiments, all numerical results are presented in Table. \ref{tab_ablation} and Table. \ref{tab_back}. 

\begin{figure}[!t]
	\centering
	\includegraphics[width=0.4\textwidth]{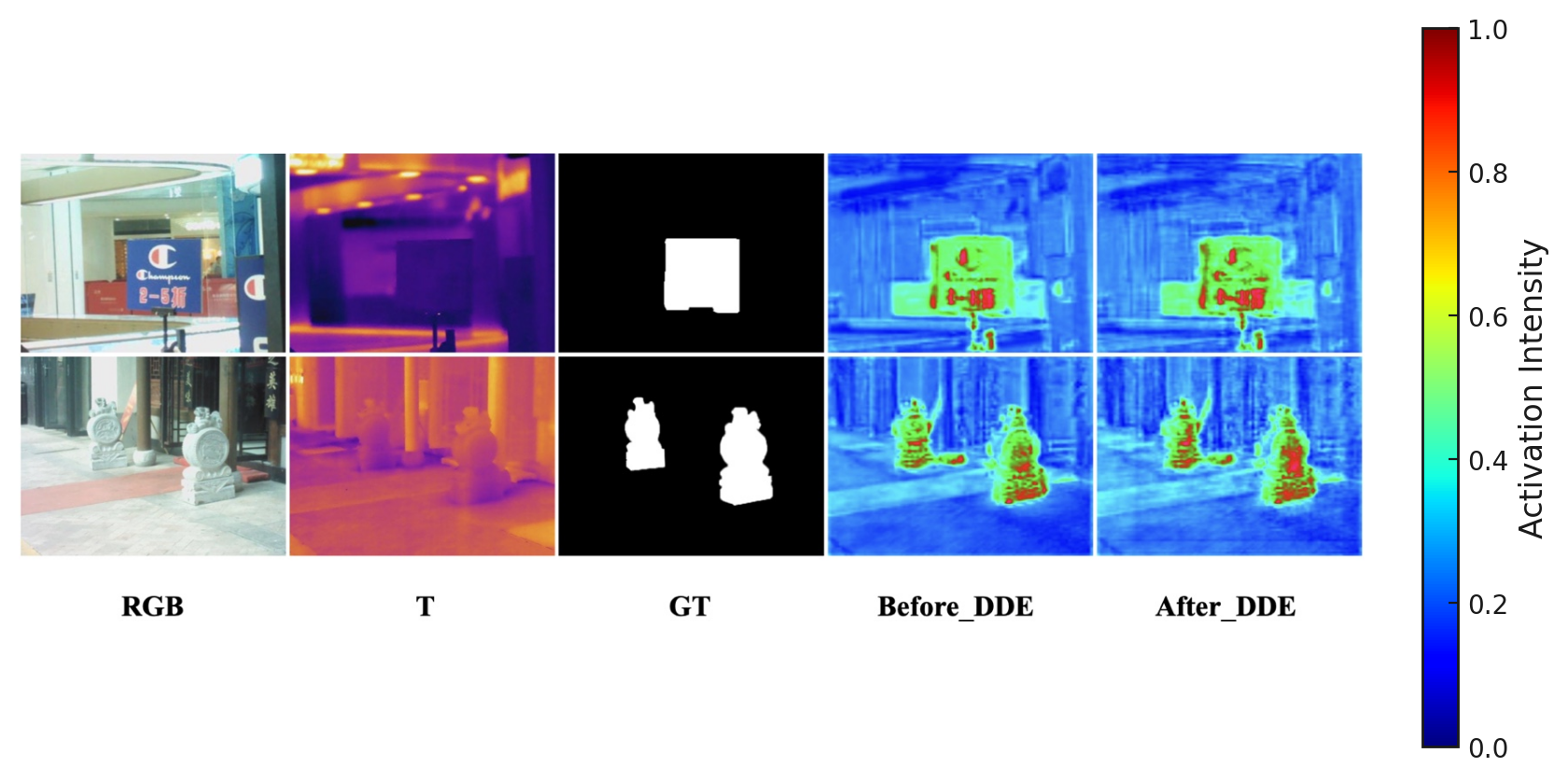}
	\caption{\small{Illustration of the impact of dense detail enhancement module.}} 
	\label{fig_hot_dde}
\end{figure}

\begin{figure}[!t]
	\centering
	\includegraphics[width=0.4\textwidth]{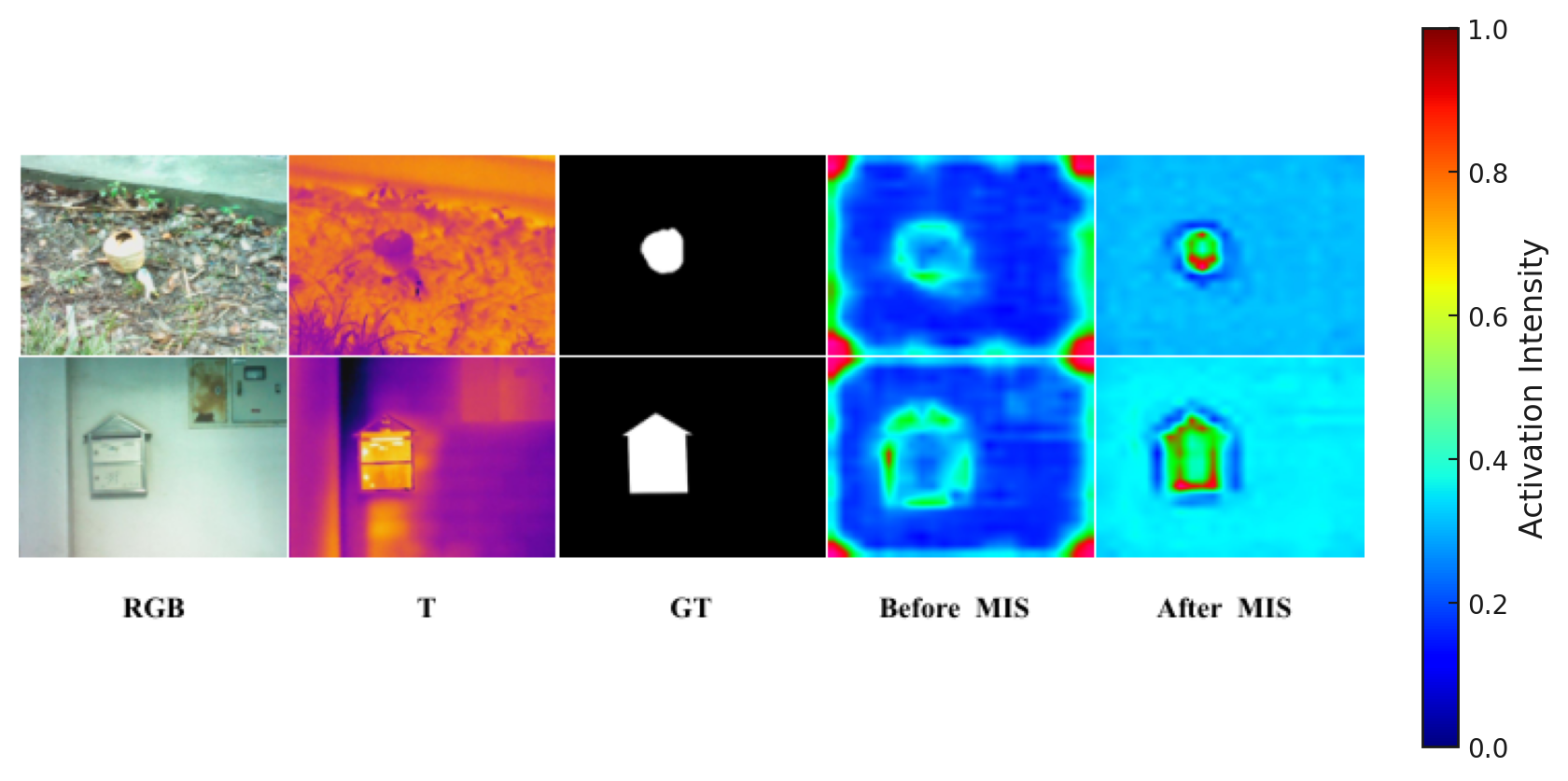}
	\caption{\small{Illustration of the impact of mutual interaction semantic module.}} 
	\label{fig_hot_mis}
\end{figure}

To validate the functionality of each module, we conduct multiple experiments, all numerical results are listed in Table. \ref{tab_ablation}. First, to demonstrate the effectiveness of the selective optimization module, we remove the region guidance stage and replace the SO module with addition, multiplication, or concatenation to integrate the RGB and thermal features, respectively. From the comparison results in Table. \ref{tab_ablation} (rows 1-3 and 10), we can see that with the absence of SO module, low-quality features are directly fused with high-quality features, leading to varying degrees of detection performance degradation. In addition, in the fourth row of the Table. \ref{tab_ablation}, we replace the selective optimization module with a Pixel-wise Soft Gating strategy, in which a sigmoid map is used to weight the RGB and thermal features at each pixel. As shown by the results, although this strategy outperforms simple addition, multiplication, or concatenation, it still leads to a noticeable performance degradation.
Then, we design two experiments (\emph{i.e.}, R$\longrightarrow$T and T$\longrightarrow$R), where the  whole region guidance stage is deleted and the two modalities features are directly fused according to Fig. \ref{fig_so} (a) or Fig. \ref{fig_so} (b). Due to the cancellation of the region guidance stage, the dual-modal feature fusion method without similarity calculation will result in low-quality feature maps being used to optimize high-quality feature maps, thereby reducing detection accuracy, which is further confirmed by the results in Table \ref{tab_ablation} (rows 5-6 and 10). After that, we successively conduct three experiments (\emph{i.e.}, $w/o $ DDE, $w/o $ MIS and $w/o $ DDE $\&$ MIS). To be specific, for the $w/o $ DDE, we remove the dense detail enhancement (DDE) module and merge the high-level features from the mutual interaction semantic (MIS) module with the low-level features from the selective optimization (SO) module in a bottom-up manner to generate the final saliency map. Besides, for the $w/o $ MIS, we remove the MIS module and similarly integrate high-level features from the SO module with the low-level features from the DDE module in a bottom-up manner. Furthermore, we remove both DDE module and MIS module in the $w/o $ DDE $\&$ MIS, and directly fuse the multi-level features generated from the SO module. From Table. \ref{tab_ablation} (rows 7-10), we observe a decline in network performance with the removal of key module.  Moreover, we visualize the feature maps before and after the application of DDE module  and MIS module, as shown in Fig. \ref{fig_hot_dde} and Fig. \ref{fig_hot_mis}. For instance, the feature map generated before the SO module primarily focuses on the approximate location of the salient object, whereas the feature map produced after the SO module accurately highlights the precise location of the object. To further evaluate the impact of different backbone, we replace the SwinTransformer \cite{liu2021swinnet} with ResNet-18\cite{he2016deep}, ResNet-34\cite{he2016deep} and ResNet-50\cite{he2016deep} as shown in Table. \ref{tab_back}. Compared to ResNet-18, our model gains improvement of $13.2\%$ in terms of $F_{\beta}$ and  $7.4\%$ in terms of $S_{\alpha}$, respectively. In addition, compared to ResNet-34 and ResNet-50, our method demonstrates varying degrees of performance improvement, indicating that SwinTransformer possesses stronger feature extraction capabilities than ResNet-18, ResNet-34 and ResNet-50. Moreover, we attempt to replace the SwinTransformer adopted in this paper with stranger feature extractors (\emph{i.e}., SAM and DINO). As shown by the results in Table. \ref{tab_back}, the model performance does not improve with the use of these large models; instead, it even degrades. This phenomenon can be attributed to the fact that frozen large-scale models are not well adapted to RGB-T imagery, and thus require dedicated adaptor to effectively exploit their representations in this task.
\subsection{Failure Cases}

\begin{figure}[!t]
	\centering
	\includegraphics[width=0.4\textwidth]{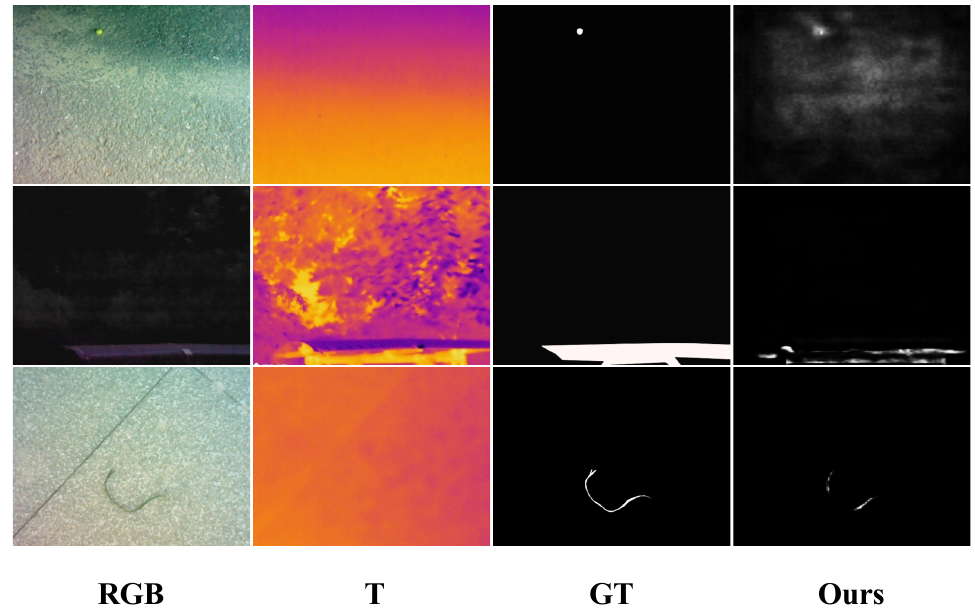}
	\caption{\small{Example of failure case.}} 
	\label{fig_faliure}
\end{figure}
To give a comprehensive presentation of our RSONet, we list some failure cases. As shown in Fig. \ref{fig_faliure} (the 1$^{st}$ and 3$^{th}$ rows),  when the salient objects in an image are extremely small or fine, our network struggles to detect them accurately. Besides, when both RGB images and thermal maps are in low quality, as shown in Fig. \ref{fig_faliure} (the 2$^{nd}$ row) , even with the use of region guidance stage and the selective optimization module, the selected feature maps still contain significant noise, severely impacting the detection results.

\section{Conclusion}
This paper proposes a novel RGB-T salient object detection network called RSONet, which includes a region guidance stage and saliency generation stage. The region guidance stage employs the encoder-decoder structure which consists of the context interaction (CI) module and spatial-aware fusion (SF) module to yield the guidance maps from RGB, thermal, and RGBT images. Then, the saliency generation stage  determines the dominant modal feature for fusion within the selective optimization (SO) module by calculating the similarity of the guidance maps. Besides, the dense detail enhancement (DDE) module and mutual interaction sementic (MIS) module are proposed to optimize the detail and location information. Overall, the proposed RSONet shows competitive  results in RGB-T salient object detection compared with other twenty-eight state-of-the-art methods. In future work, we plan to explore more effective strategies for capturing and leveraging the complementary information between RGB and thermal modalities, further extending the applicability and robustness of our approach. 

\bibliographystyle{IEEEtran}
\bibliography{reference}
\end{document}